\documentclass[a4paper, 12pt]{iopart}
\usepackage{cite}
\usepackage{color,soul}
\usepackage{dcolumn}
\usepackage{bm}
\usepackage{comment}
\usepackage[usenames,dvipsnames]{xcolor}
\usepackage[colorlinks=true,citecolor=Blue,linkcolor=RubineRed,urlcolor=Blue]{hyperref}
\usepackage{graphicx}
\usepackage[font=small]{caption}
\usepackage{iopams}

\usepackage{adjustbox}

\usepackage{algorithm}
\usepackage{algpseudocode}
\usepackage{booktabs}

\newcommand{\argmin}{\mathop{\mathrm{arg\,min}}}

\begin{document}

\title{Optimised Feature Subset Selection via Simulated Annealing}

\author{F. Mart\'{i}nez-Garc\'{i}a$^1$, A. Rubio-Garc\'{i}a$^2$, S. Fern\'{a}ndez-Lorenzo$^2$, J. J. Garc\'{i}a-Ripoll$^1$, D. Porras$^1$}
\address{$^1$ Instituto de Física Fundamental, IFF-CSIC, Calle Serrano 113b, Madrid 28006, Spain}
\address{$^2$ Inspiration-Q, Calle Tablas de Daimiel, 7, 2, Madrid, 28924, Spain}
\ead{f.martinez@iff.csic.es}
\vspace{10pt}
\begin{indented}
\item[]\today
\end{indented}

\begin{abstract}
    We introduce SA-FDR, a novel algorithm for $\ell_0$-norm feature selection that considers this task as a combinatorial optimisation problem and solves it by using simulated annealing to perform a global search over the space of feature subsets. The optimisation is guided by the Fisher discriminant ratio, which we use as a computationally efficient proxy for model quality in classification tasks. Our experiments, conducted on datasets with up to hundreds of thousands of samples and hundreds of features, demonstrate that SA-FDR consistently selects more compact feature subsets while achieving a high predictive accuracy. This ability to recover informative yet minimal sets of features stems from its capacity to capture inter-feature dependencies often missed by greedy optimisation approaches. As a result, SA-FDR provides a flexible and effective solution for designing interpretable models in high-dimensional settings, particularly when model sparsity, interpretability, and performance are crucial.
\end{abstract}

\section{Introduction}

Advances in data acquisition, storage, and processing techniques have resulted in the generation of vast and increasingly complex datasets. This abundance of data offers new opportunities to improve the performance of classification or forecasting models. However, this abundance also presents significant challenges, particularly in terms of dimensionality and computational efficiency. Moreover, as datasets grow in size and complexity, they often include a large number of variables, many of which may be irrelevant, redundant, or subject to noise. Including such variables in a model can obscure underlying relationships, lead to overfitting, and compromise the stability and interpretability of the model~\cite{hastie2009elements, kuhn2013applied}. A fundamental step to alleviate these problems consists of selecting the most relevant features required to build efficient and interpretable models. By eliminating superfluous variables, feature selection helps simplify models, reducing training time and improving generalisation on unseen data~\cite{guyon2003introduction, guyon2008feature, chandrashekar2014survey}. This is particularly important in high-dimensional settings, such as those encountered in genomics~\cite{guyon2002gene, saeys2007review}, and text or image analysis~\cite{forman2003extensive, wright2008robust}, where the number of variables can vastly exceed the number of observations, as well as in datasets with considerable noise, such as those in finance~\cite{tsay2005analysis}, sensor measurements~\cite{luo2002multisensor}, or medical records~\cite{yang2024addressing}.

Despite its importance, feature selection is a challenging task, since the number of possible feature subsets grows exponentially with the total number of features~\cite{bertsimas2016best, sato2016feature, bertsimas2021sparse}. This makes brute force search of optimal models computationally prohibitive even for moderately sized datasets. Assessing the quality of a subset with respect to model accuracy and generalizability is also inherently difficult, particularly in the presence of feature correlations, noise, or limited sample sizes~\cite{varma2006bias, cawley2010over}. Existing methods are typically categorised as filter, wrapper, or embedded approaches~\cite{guyon2003introduction, liu2007computational, cai2018feature, moslemi2023tutorial}. Filter methods rely on statistical criteria computed independently of any model, offering efficiency but potentially overlooking the effect of feature correlations. Embedded methods integrate feature selection into the model training process by applying constraints or regularisation, but remain dependent on the structure and assumptions of the chosen model. Finally, wrapper methods train and validate a predictive model for different candidate subsets, yielding better results in exchange for an increased computational cost. Among these approaches, wrapper methods are particularly well-suited to incorporate combinatorial optimisation techniques, as they directly evaluate the performance of candidate feature subsets. This opens the door to sophisticated search strategies that explore the combinatorial nature of feature selection more effectively.

Combinatorial optimisation problems have long been a subject of interest in the field of statistical physics~\cite{edwards1975theory, barahona1982computational, mezard1987spin}, which studies complex systems composed of many interacting components. This research has led to the development of computational methods to solve difficult optimisation problems, such as simulated annealing~\cite{kirkpatrick1983optimization}. Simulated annealing is an algorithm inspired by the thermodynamic process of slowly cooling a material to reach a low-energy state. By allowing probabilistic transitions that occasionally accept worse solutions --- depending on a simulated temperature --- the algorithm avoids becoming trapped in local minima and can converge toward a global minimum as the temperature gradually approaches zero. The success of algorithms based on simulated annealing for studying discrete and high-dimensional combinatorial optimisation problems~\cite{wang2015comparing, martinez2025problem, rubio2022portfolio, rubio2024accurate} makes it a promising candidate for approaching feature selection problems.

In this work, we propose a wrapper method based on the simulated annealing algorithm for the feature selection problem. Specifically, we use it to find feature subset candidates for classification tasks, assuming a logistic regression model~\cite{hosmer2013applied}. Moreover, we consider the Fisher Discrimination Ratio (FDR)~\cite{fisher1936use, theodoridis2015machine} as a proxy of the logistic regression quality, which allows us to avoid the calculation of several cross-entropy values, resulting in a reduction of the time consumed by the algorithm. We test our approach and compare it with other widely used algorithms for feature selection and obtain high-quality results for different datasets, which allow us to obtain models that consider few features while achieving high accuracies.

This work is organised as follows: In Section~\ref{sec:classification_model}, we introduce the basic concepts related to the logistic model that we will consider, as well as the Fisher Discriminant Ratio, which we will use as a proxy for the quality of our selected features. In Section~\ref{sec:feature_selection}, we introduce the simulated annealing algorithm and how we use it to solve the combinatorial optimisation problem of feature selection. We will also explain the details of our benchmark simulations and the algorithms that we use to compare our proposed algorithm. After introducing these concepts, in Section~\ref{sec:results} we show the results obtained for a wide variety of publicly available datasets. Finally, in Section~\ref{sec:conclusions} we present our conclusions to this study.

\section{Classification Model}
\label{sec:classification_model}

Consider a dataset composed of N samples, $\{\bm{x}^{(i)}, y^{(i)}\}^N_{i=1}$, where the vector $\bm{x}^{(i)} \in \mathbb{R}^K$ corresponding to the $i$-th sample contains a set of $K$ feature values, and $y^{(i)}$ is a binary variable that associates the sample with one of two possible classes, $\mathcal{C}_0$ or $\mathcal{C}_1$. While we will focus on this binary classification case, we note that the methods shown in this work are generalizable to cases with multiple classes. The goal of feature selection is to select a subset of $k<K$ features that are most relevant to predicting the class of each sample. To do this, we need to define the underlying model that will be used to perform the classification given a subset of features. In this work, we will consider logistic models.

\subsection{Logistic Regression}

A logistic regression is a classification model that assigns to each sample $\bm{x}$ a probability of belonging to a given class. For our binary classification case, we will use $p(\bm{x})$ for the probability of $\mathcal{C}_0$ and $1 - p(\bm{x})$ for the probability of $\mathcal{C}_1$. The log-odds are given by a linear relation:
\begin{equation}
    \log \frac{p(\bm{x})}{1-p(\bm{x})} = \beta_0 + \sum^k_{j=1}\beta_j x_j,
\end{equation}
with the probability $p(\bm{x})$ being given by:
\begin{equation}
    p(\bm{x})=\frac{1}{1+\exp{\left[-\beta_0 - \sum^k_{j=1}\beta_j x_j\right]}},
\end{equation}
where $\bm{\beta}=(\beta_0, \beta_1, ..., \beta_k)$ are the parameters of the model, with the last $k$ parameters being related to the chosen subset of explanatory features, and $\beta_0$ being an intercept term. We may obtain the optimal regression parameters by minimising the cross-entropy over the classification dataset:
\begin{equation}
    \bm{\beta}=\argmin_{\beta_0, ..., \beta_k} \sum^N_{i=1}-\left[ y^{(i)} \log(p(\bm{x}^{(i)}) + (1-y^{(i)}) \log(1 - p(\bm{x}^{(i)}))\right].
\end{equation}
While there is no closed-form solution for the optimal regression parameters, we can find them using standard unconstrained optimisation algorithms used in machine learning, such as gradient descent or the BFGS method~\cite{nocedal1999numerical, fletcher2000practical}. We use the latter method in this work.

Our goal is to obtain a feature subset of a given size $k$ that results in the best classification when considering a logistic model. While we can train a logistic model in a reasonable time, our proposed feature selection algorithm will need to evaluate a considerable number of models to find an optimised feature subset. Due to this, we will consider the Fisher Discriminant Ratio as a faster alternative that serves as a proxy to estimate the quality of a given feature subset for linear classification.

\subsection{Fisher Linear Discriminant}
\label{sec:FLD}

The Fisher Linear Discriminant~\cite{fisher1936use, theodoridis2015machine} is a linear projection method that, given a set of features, finds their best linear combination that separates between two or more classes. To do this, it provides a projection direction $\mathbf{w} \in \mathbb{R}^k$ that satisfies a compromise between maximising the between-class variance and minimising the within-class variance. This defines the \textit{Fisher Discriminant Ratio} (FDR), which is given by:
\begin{equation}
    \mbox{FDR}=\frac{(\mu'_1 - \mu'_0)^2}{\sigma^2_1 + \sigma^2_0},
\end{equation}
where $\mu'_i$ and $\sigma^2_i$ with $i=0,1$ are the mean value and the variance of the projections of the elements of $\mathcal{C}_i$ over $\mathbf{w}$, respectively. Of course, these values --- and therefore the FDR value --- depend on the choice of $\mathbf{w}$. This vector is the one that maximises the FDR value, optimising the compromise between within-class variance and between-class variance. To find the vector $\mathbf{w}$, we can consider that:
\begin{equation}
    (\mu'_1 - \mu'_0)^2 = \mathbf{w}^T (\boldsymbol{\mu_1} - \boldsymbol{\mu_0})(\boldsymbol{\mu_1} - \boldsymbol{\mu_0})^T \mathbf{w} \equiv \mathbf{w}^T S_B \mathbf{w},
\end{equation}
where $\boldsymbol{\mu_i} \in \mathbb{R}^k$ is the position of the class means in feature space. Similarly, we have:
\begin{equation}
    \sigma^2_i = \mathbb{E}\left[\mathbf{w}^T(\bm{x}_{\mathcal{C}_i} - \boldsymbol{\mu_i})(\bm{x}_{\mathcal{C}_i} - \boldsymbol{\mu_i})^T\mathbf{w}\right] \equiv \mathbf{w}^T \Sigma_i \mathbf{w},
\end{equation}
where $\bm{x}_{\mathcal{C}_i}$ means that we only consider the feature vectors associated with class $\mathcal{C}_i$, and $\Sigma_i$ with $i=0,1$ are the within-class variances. Given these definitions, the FDR value can be written as:
\begin{equation}
    \mbox{FDR}=\frac{\mathbf{w}^T S_B \mathbf{w}}{\mathbf{w}^T S_W \mathbf{w}},
\end{equation}
where $S_W \equiv \Sigma_0 + \Sigma_1$. This is a generalised Rayleigh quotient~\cite{parlett1998symmetric, horn2012matrix} and its maximum value is achieved when:
\begin{equation}
    \mathbf{w} \propto S^{-1}_W (\boldsymbol{\mu_1} - \boldsymbol{\mu_0}),
\end{equation}
with a FDR value given by:
\begin{equation}
    \mbox{FDR} = (\boldsymbol{\mu_1} - \boldsymbol{\mu_0})^T S^{-1}_W (\boldsymbol{\mu_1} - \boldsymbol{\mu_0}).
\end{equation}

Applying this method to a set of $k$ selected features results in an FDR value that measures the quality of these features to perform a linear separation of the two possible classes. This method can then be used as a metric of the quality of a subset of features for classification. While the FDR criterion is derived from a different framework than logistic regression, both methods seek to separate classes based on a linear combination of features. Although FDR does not capture the same aspects as logistic regression --- such as probability estimation --- it offers a fast measure of class separability that correlates well with the cross entropy of a logistic regression (see Fig.~\ref{fig:FDR_full_figure}). This makes the FDR criterion a great fast-to-calculate proxy metric to substitute the cross-entropy during the combinatorial optimisation process that we introduce in the following section.

We have defined metrics to quantify the capability of a subset of $k$ features to estimate the correct class of a dataset element. However, we still need a method to find the best possible feature subset from all the possible combinations. Therefore, we are presented with a combinatorial optimisation problem that we propose to solve with the Simulated Annealing algorithm.

\section{Feature Selection}
\label{sec:feature_selection}

In this section, we describe how we use Simulated Annealing to find the optimal set of $k$ features to classify a given dataset. We then explain how we use this algorithm with a cross-validation approach to select the optimal value of $k$.

\subsection{Simulated Annealing}

\begin{figure}[t]
\centering
\includegraphics[width=\textwidth]{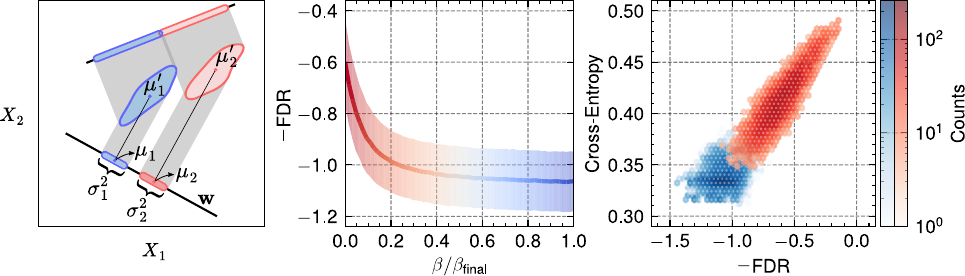}
\caption{Left: Example of the Fisher Linear Discriminant method to find a line that optimally separates two classes, $1$ and $2$. The found line, characterised by $\mathbf{w}$, maximises the distance between the projections of the classes, while minimising the projected within-class variance. The top line shows a comparison when projecting over a random line. Middle: Minimisation of $-$FDR using the SA algorithm by increasing the inverse temperature for the UCI SPECTF Heart dataset, considering $k=5$ features. The continuous line represents the mean of all the replicas, and the shaded area the standard deviation. Right: Densities of the $-$FDR and cross-entropy values for the replicas at infinite temperature (red) and after cooling with SA (blue) for the same dataset and $k$ value, showcasing the initial and final states of the middle figure. SA allows us to sample feature subsets of high FDR, which are correlated to a low cross-entropy, resulting in optimal feature subsets for the logistic regression.}
\label{fig:FDR_full_figure}
\end{figure}

Simulated Annealing (SA) is a Monte Carlo algorithm that can be used to sample a target distribution. In our case of interest, this will be the Boltzmann distribution. This algorithm considers a set of $R$ \textit{replicas} $\mathbf{s}^{(r)}$, $r=1,..., R$, with each replica being a subset of $k$ features with an associated objective function or \textit{energy}. While the term energy in this algorithm comes from the physical origins of this method, this term is interchangeable with any scalar objective function to be minimised. In our case, this is the FDR value (with a negative sign, since we want to maximise it). When the features of these replicas are initialised randomly, they can be considered as samples from a Boltzmann distribution at a temperature $T_0\rightarrow \infty$ or inverse temperature $\beta_0=1/T_0\rightarrow 0$. Then, for each replica $\mathbf{s}$ with FDR value $\mathrm{FDR}(\mathbf{s})$, it proposes a change in their feature subset, resulting in $\mathbf{s}^\prime$, and computes the corresponding $\mathrm{FDR}(\mathbf{s}^\prime)$ value. This change is then accepted (or rejected) by considering the Metropolis rule~\cite{robert1999monte}:
\begin{equation}
\label{eq:MH_rule}
P(\mathbf{s'}|\mathbf{s}) = \min \left(
1 ,\,e^{\beta_1\left[\mathrm{FDR}(\mathbf{s}')-\mathrm{FDR}(\mathbf{s})\right]} 
\right),
\end{equation}
where $\beta_1>\beta_0$. After proposing enough changes, the initial replicas obtained from a thermal distribution at inverse temperature $\beta_0$ are transformed into approximate samples at inverse temperature $\beta_1$. For each temperature value, we perform a total of $N_S$ sweeps, consisting of $K$ swaps of a feature considered in the subset with another outside of it. The samples obtained can then be used to obtain samples at a higher value of $\beta$. This defines an iterative process that anneals the replicas to increasing values $\beta_t$, with $t=1,..., N_T$, until reaching a target final temperature (see Fig.~\ref{fig:FDR_full_figure}). If the schedule by which we decrease the temperature is slow enough, the SA algorithm is guaranteed to sample the target distribution \cite{mitra1986convergence}. These steps constitute the SA algorithm, which finds a set of replicas with FDR values close to the optimal if a high enough final temperature and enough computational resources are considered.

\begin{algorithm}[t]
\caption{Simulated Annealing for Feature Selection}
\begin{algorithmic}[1]
\Require Dataset, sweeps ($N_S$), feature subset size ($k$), temperature steps ($\beta_{\mbox{steps}}$), and temperature step size ($\epsilon$).
\State Initialize replicas $\mathbf{s}^{(r)}$ and compute the values $\mbox{FDR}(\mathbf{s}^{(r)})$. Set $\beta_0=0$.
\State Set $\sigma$ as the standard deviation of $\mbox{FDR}(\mathbf{s}^{(r)})$.
\State Calculate different $S_B$ and $S_W$ matrices using different batches of the dataset.
\For{$t=0$ to $\beta_{\mbox{steps}} - 1$}
    \For{$\mathbf{s}$ in $\{\mathbf{s}^{(r)}\}^R_{r=1}$}
        \For{$i=1$ to $N_S \cdot K$}
            \State Generate new candidate $\mathbf{s}'$. \State Compute $\Delta = \mbox{FDR}(\mathbf{s}) - \mbox{FDR}(\mathbf{s}')$ using a random batch matrix set.
            \State \textbf{if} $\Delta < 0$ \textbf{then} Accept $\mathbf{s'}$ \textbf{else} Accept $\mathbf{s'}$ with probability $e^{-\beta_t \Delta}$.
\EndFor
\EndFor
\State \textbf{if} the mean of $\mbox{FDR}(\mathbf{s})$ has converged \textbf{then} break
\State Set new inverse temperature $\beta_{t+1} \rightarrow \beta_t + \epsilon/\sigma$.
\EndFor
\State Calculate logistic regressions with the feature subsets of each replica.
\State \Return Best feature subset, $\mathbf{s}^*$, that minimizes the cross-entropy
\end{algorithmic}
\label{alg:simulated_annealing}
\end{algorithm}

The SA algorithm described can be used to find the subset of $k$ features with maximum FDR value. However, as explained in Sec.~\ref{sec:FLD}, we consider the FDR value as an efficient proxy of the cross-entropy, since both are related. A consequence is that the feature subset that maximises the FDR value might differ slightly from the one that minimises the cross-entropy. To solve this problem, we evaluate the cross-entropies of the different resulting replicas at the end of the annealing and select the one that minimises it. Additionally, to keep a baseline noise that avoids overfitting and helps produce replicas that have not converged to the same FDR local maximum, our implementation of the algorithm calculates a set of matrices $S_B$ and $S_W$ (defined in Sec.~\ref{sec:FLD}) using different batches of the dataset. For each step and replica of the SA algorithm, we select randomly one set of these batch matrices and use them to calculate the FDR value after the proposed change.

Finally, we use a temperature schedule such that each step in the inverse temperature is proportional to the inverse of the FDR standard deviation at $\beta=0$. Our algorithm runs until it reaches a target number of temperature steps (in our case 100) or until the FDR mean value converges. We provide a pseudocode explanation of our proposed algorithm in Algorithm~\ref{alg:simulated_annealing}.

\subsection{Cross-validation}

To obtain a metric of the quality of our solutions, we first need to estimate the optimal feature number $k^*$ for a dataset. To do this, we follow a cross-validation approach~\cite{james2013introduction}, where we split the dataset into a train and test set, each with 80\% and 20\% of the samples from the original dataset, respectively. We maintain the original dataset class balance in the train and test sets. For each value of $k$, we do 5 splits -- also called folds -- of the train set into a train and a validation set for each fold in a 75\% to 25\% relation, respectively. Then, for each fold, we run our proposed feature selection algorithm in the train set and measure the Receiving Operator Characteristic-Area Under the Curve (AUC)~\cite{fawcett2006introduction} of the resulting model in the validation set. We then calculate the mean AUC value in the validation set using the results from all the folds and find the optimal value $k^*$, which we define as the minimum value of $k$ with an AUC value that is one standard deviation below the maximum value found. We then use our algorithm over the combined train and validation set considering $k^*$ features, and use the resulting model on the test set to obtain the corresponding AUC value. We repeat this process up to 20 times to obtain an average value of the AUC and $k^*$.

\begin{figure}[t]
\centering
\includegraphics[width=0.95\textwidth]{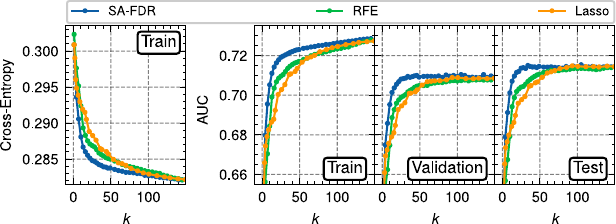}
\caption{Results of the cross-entropy (on train dataset) and AUC scores (on train, validation, and test datasets) obtained by using the SA-FDR, RFE, and Lasso feature selection algorithms on the Loan Default dataset. The features selected by SA-FDR result in cross-entropy values that improve those obtained by the other two algorithms. This results in a better logistic model that achieves better AUC scores. The train and validation scans are repeated a total of five folds. The obtained mean validation curve is used to estimate the optimal number of features, $k^*$, and the resulting model is used on the test dataset. This process is repeated 20 times to obtain the averages of $k^*$ and AUC for each algorithm, shown in Table~\ref{tab:results}.}
\label{fig:cross_validation}
\end{figure}

To compare our algorithm with other widely used feature selection algorithms, we also run the above cross-validation process for obtaining the optimal $k^*$ and test set AUC score, but replacing the feature selection using SA with the Recursive Feature Elimination (RFE)~\cite{guyon2002gene} and Lasso algorithms~\cite{tibshirani1996regression, hastie2009elements}, which are available within the Scikit-learn library~\cite{scikit-learn}. Both of these algorithms, as well as our approach, find a feature subset candidate and train a logistic model with it. An example of this benchmark process is shown in Fig.~\ref{fig:cross_validation}. While for both our method and the RFE, one can set a target value of $k$, for Lasso, the control parameter is a regularisation strength, $C$. Therefore, for the Lasso results, we scan over different values of $C$, find the optimal value, $C^*$, and calculate the corresponding average value of selected features as the value $k^*$.

\section{Results}
\label{sec:results}

We consider different datasets with a diverse number of samples and features that are publicly available from the UCI Machine Learning repository~\cite{kelly2023uci}. The dataset names and the abbreviation that we use to refer to them are as follows: SPECTF Heart (Heart)~\cite{cios2001spectf}, Breast Cancer Wisconsin Diagnostic (Cancer)~\cite{wolberg1993breast}, Ozone Level Detection Eight Hours (Ozone 8)~\cite{zhang2008ozone}, Spambase (Spambase)~\cite{hopkins1999spambase}, Predict Students' Dropout and Academic Success (Student)~\cite{realinho2021predict}, 
Taiwanese Bankruptcy Prediction (Bankruptcy)~\cite{uci2020taiwanese}, Madelon (Madelon)~\cite{guyon2004madelon}, Default of Credit Card Clients (Card Default)~\cite{yeh2009default}. We also consider the publicly available Loan Default dataset~\cite{yasser2022loan}, which we clean from samples with missing values and highly correlated features. We also calculate the mean and standard deviation of each feature from the test dataset and use these values to standardise the entire dataset. Specifically, for each feature, we subtract the corresponding mean and divide by the corresponding standard deviation from the test set.

\begin{table}[t]
\centering
\caption{Performance comparison of our proposed algorithm, SA-FDR, with RFE and Lasso over datasets with different number of samples, $N$, and total features, $K$. We show the results of the average $k^*$ and AUC found. The best values found across the three algorithms are marked in bold. We also provide the time required for each algorithm to evaluate a fold. We note that the times for the Loan Default dataset increase considerably since we are considering values of $k$ up to 150. However, for SA-FDR we find that considering up to $k=30$ might have been enough, which would drastically reduce the time required compared to the other two algorithms.}
\label{tab:results}
\begin{adjustbox}{width=\textwidth}
\begin{tabular}{lcc|ccc|ccc|ccc}
\toprule
\textbf{Dataset} &  &  & \multicolumn{3}{c|}{\textbf{SA-FDR}} & \multicolumn{3}{c|}{\textbf{RFE}} & \multicolumn{3}{c}{\textbf{Lasso}} \\
                 & \textbf{$N$} & \textbf{$K$} & $k^*$ & AUC & t(s) & $k^*$ & AUC & t(s) & $k^*$ & AUC & t(s) \\
\midrule
Heart     & 267 & 44   & \textbf{3.40} & 0.8165 & 4.3 & 3.65 & 0.8106 & 0.7 & 6.08 & \textbf{0.8211} & 0.9 \\
Cancer     & 569 & 30   & \textbf{3.60} & 0.9922 & 2.2 & 5.20 & 0.9921 & 0.8 & 5.59 & \textbf{0.9933} & 0.9 \\
Ozone 8    & 1847 & 72      & \textbf{9.35} & \textbf{0.9058} & 5.2 & 11.95 & 0.9020 & 1.2 & 19.59 & 0.9034 & 1.7 \\
Spambase    & 4601 & 57      & 23.35 & \textbf{0.9688} & 6.1 & \textbf{22.25} & 0.9674 & 1.5 & 37.59 & 0.9707 & 2.1 \\
Student     & 4424 & 36  & \textbf{10.05} & 0.9162 & 3.9 & 14.70 & 0.9165 & 0.7 & 13.88 & \textbf{0.9170} & 1.4 \\
Bankruptcy   & 6819 & 95       & \textbf{2.45} & \textbf{0.9281} & 10.9 & 3.95 & 0.9161 & 2.6 & 6.34 & 0.9270 & 6.1 \\
Madelon    & 20000 & 500      & \textbf{1.60} & \textbf{0.6339} & 250 & 6.00 & 0.6250 & 5.2 & 2.60 & 0.6329 & 9.8 \\
Card Default   & 30000 & 23       & \textbf{10.00} & 0.7214 & 10.8 & 10.75 & 0.7215 & 1.5 & 13.25 & \textbf{0.7220} & 8.6 \\
Loan Default   & 103302 & 397        & \textbf{26.80} & \textbf{0.7120} & 5600 & 96.00 & 0.7101 & 2300 & 70.36 & 0.7112 & 1100 \\
\bottomrule
\end{tabular}
\end{adjustbox}
\end{table}

We use our proposed algorithm by considering 50 replicas, $N_S=0.5$, $\epsilon=0.7$, and scanning for values of $k$ up to $k_{max}=30$, which is enough to find the value $k^*$ for all the datasets. For the study of the Loan Default dataset, we show the results with up to $k_{max}=150$ in Fig.~\ref{fig:cross_validation}. We use the previous datasets to obtain benchmarks of our proposed SA-FDR algorithm and compare these results with the RFE and Lasso algorithms. The results of this benchmarking process for each considered dataset, as well as the dimensions of each dataset, are shown in Table~\ref{tab:results}. From these results, we observe that SA-FDR obtains the smallest value of $k^*$ for most of the cases. Moreover, it also achieves the best AUC value for some of these datasets, and for the cases where it does not reach the best value, it finds a value that is considerably close to it. Specifically, our algorithm demonstrates a clear advantage in identifying sparse models with few features, as shown by its performance relative to the other two methods. This is particularly evident when compared to the RFE algorithm, which follows a greedy strategy by iteratively removing the feature that contributes the least to logistic regression. In contrast, our SA-FDR algorithm performs a combinatorial optimisation over each value of $k$, allowing it to explore a wider range of feature subsets. As a result, it can consider features that RFE may have mistakenly discarded early in the process, leading to more effective subset selection and improved performance, as can also be seen in the results shown in Fig.~\ref{fig:cross_validation}. Finally, we also show the time each algorithm requires to evaluate a single fold. The simulations were performed using an AMD Ryzen 9 5950x 3.4GHz and parallelising over each fold. While our proposed algorithm is the slowest of the tested algorithms, the runtime for SA-FDR is still well within the time scales that make the algorithm useful in practice. Moreover, we did not search for the hyperparameters that optimise the time; therefore, decreasing the number of replicas or sweep number might keep the same result quality while considerably reducing the algorithm runtime. Overall, these numerical results highlight the potential of SA-FDR as a robust tool for high-quality sparse model selection across a wide range of datasets of varying sizes, which can be encountered in real-world problems.

\section{Conclusions}
\label{sec:conclusions}

In this work, we proposed SA-FDR, a novel algorithm that considers the feature selection task as a combinatorial optimisation problem. It leverages simulated annealing to search the space of feature subsets, using the Fisher discriminant ratio as an objective function approximating logistic model quality. We evaluated SA-FDR on datasets with varying numbers of samples and features, with models with up to hundreds of thousands of samples and hundreds of features. Compared to standard feature selection algorithms, SA-FDR consistently identified sparser feature subsets while maintaining high predictive performance. Its ability to explore a diverse range of feature combinations allows it to capture interactions that need to be considered for obtaining sparser models, and that greedy approaches often ignore. This makes SA-FDR a valuable tool for applications where feature sparsity, model interpretability, and accuracy are essential.

We note that, although our proposed algorithm uses the Fisher discriminant ratio as the objective function and is evaluated in the context of logistic models, the underlying framework is not limited to these choices. In principle, any metric that measures model quality can serve as the objective function for the simulated annealing algorithm. Furthermore, the algorithm is not necessarily restricted to logistic regressions and can be adapted to work with other types of classifiers, including non-linear models like neural networks, as long as the objective function is correlated to the accuracy of the model considered. This flexibility opens the door for applying simulated annealing to a wide range of learning paradigms, enabling principled sparse feature selection even in more complex modelling scenarios, and possibly with performances that improve over the results shown in this work.

In summary, SA-FDR or, more generally, simulated annealing, is a valuable tool for feature selection, readily applicable to many real-world problems. By enabling richer exploration of the feature space, our approach opens the door to optimised model design and improved interpretability. Its flexibility allows it to be suitable for diverse high-dimensional datasets. Moreover, the algorithmic framework invites further refinements, such as improvements to the simulated annealing that have been proposed and used in fields such as statistical physics~\cite{wang2015comparing, miasojedow2013adaptive, barash2017gpu}. These enhancements could improve both computational efficiency and selection accuracy. Overall, our proposed framework not only contributes to current state-of-the-art methods for feature selection but also offers a versatile foundation for future developments in machine learning workflows.

\section{Acknowledgments}
\label{sec:acknowledgments}
This project has been supported by Spanish Project No. PDC2022-133486-I00, funded by MCIN/AEI/10.13039/501100011033 and by the European Union “NextGenerationEU”/PRTR”1; and CSIC Interdisciplinary Thematic Platform (PTI) Quantum Technologies (PTI-QTEP).

\section*{References}
\bibliography{sample}

\providecommand{\newblock}{}
\begin{thebibliography}{10}
\expandafter\ifx\csname url\endcsname\relax
  \def\url#1{{\tt #1}}\fi
\expandafter\ifx\csname urlprefix\endcsname\relax\def\urlprefix{URL }\fi
\providecommand{\eprint}[2][]{\url{#2}}

\bibitem{hastie2009elements}
Hastie T 2009 The elements of statistical learning: data mining, inference, and prediction

\bibitem{kuhn2013applied}
Kuhn M, Johnson K {\em et~al.\/} 2013 {\em Applied predictive modeling\/} vol~26 (Springer)

\bibitem{guyon2003introduction}
Guyon I and Elisseeff A 2003 {\em Journal of machine learning research\/} {\bf 3} 1157--1182

\bibitem{guyon2008feature}
Guyon I, Gunn S, Nikravesh M and Zadeh L~A 2008 {\em Feature extraction: foundations and applications\/} vol 207 (Springer)

\bibitem{chandrashekar2014survey}
Chandrashekar G and Sahin F 2014 {\em Computers \& electrical engineering\/} {\bf 40} 16--28

\bibitem{guyon2002gene}
Guyon I, Weston J, Barnhill S and Vapnik V 2002 {\em Machine learning\/} {\bf 46} 389--422

\bibitem{saeys2007review}
Saeys Y, Inza I and Larranaga P 2007 {\em Bioinformatics\/} {\bf 23} 2507--2517

\bibitem{forman2003extensive}
Forman G 2003 {\em Journal of machine learning research\/} {\bf 3} 1289--1305

\bibitem{wright2008robust}
Wright J, Yang A~Y, Ganesh A, Sastry S~S and Ma Y 2008 {\em IEEE transactions on pattern analysis and machine intelligence\/} {\bf 31} 210--227

\bibitem{tsay2005analysis}
Tsay R~S 2005 {\em Analysis of financial time series\/} (John wiley \& sons)

\bibitem{luo2002multisensor}
Luo R~C and Kay M~G 2002 {\em IEEE Transactions on Systems, Man, and Cybernetics\/} {\bf 19} 901--931

\bibitem{yang2024addressing}
Yang J, Triendl H, Soltan A~A, Prakash M and Clifton D~A 2024 {\em BMC Medical Informatics and Decision Making\/} {\bf 24} 183

\bibitem{bertsimas2016best}
Bertsimas D, King A and Mazumder R 2016 {\em The annals of statistics\/}  813--852

\bibitem{sato2016feature}
Sato T, Takano Y, Miyashiro R and Yoshise A 2016 {\em Computational Optimization and Applications\/} {\bf 64} 865--880

\bibitem{bertsimas2021sparse}
Bertsimas D, Pauphilet J and Van~Parys B 2021 {\em Machine Learning\/} {\bf 110} 3177--3209

\bibitem{varma2006bias}
Varma S and Simon R 2006 {\em BMC bioinformatics\/} {\bf 7} 1--8

\bibitem{cawley2010over}
Cawley G~C and Talbot N~L 2010 {\em The Journal of Machine Learning Research\/} {\bf 11} 2079--2107

\bibitem{liu2007computational}
Liu H and Motoda H 2007 {\em Computational methods of feature selection\/} (CRC press)

\bibitem{cai2018feature}
Cai J, Luo J, Wang S and Yang S 2018 {\em Neurocomputing\/} {\bf 300} 70--79

\bibitem{moslemi2023tutorial}
Moslemi A 2023 {\em Engineering Applications of Artificial Intelligence\/} {\bf 126} 107136

\bibitem{edwards1975theory}
Edwards S~F and Anderson P~W 1975 {\em Journal of Physics F: Metal Physics\/} {\bf 5} 965

\bibitem{barahona1982computational}
Barahona F 1982 {\em Journal of Physics A: Mathematical and General\/} {\bf 15} 3241

\bibitem{mezard1987spin}
M{\'e}zard M, Parisi G and Virasoro M~A 1987 {\em {Spin glass theory and beyond: An Introduction to the Replica Method and Its Applications}\/} vol~9 (World Scientific Publishing Company)

\bibitem{kirkpatrick1983optimization}
Kirkpatrick S, Gelatt C~D and Vecchi M~P 1983 {\em Science\/} {\bf 220} 671--680 publisher: American Association for the Advancement of Science

\bibitem{wang2015comparing}
Wang W, Machta J and Katzgraber H~G 2015 {\em Physical Review E\/} {\bf 92} 013303

\bibitem{martinez2025problem}
Mart{\'\i}nez-Garc{\'\i}a F and Porras D 2025 {\em arXiv preprint arXiv:2501.07638\/}

\bibitem{rubio2022portfolio}
Rubio-Garc{\'\i}a {\'A}, Garc{\'\i}a-Ripoll J~J and Porras D 2022 {\em arXiv preprint arXiv:2210.00807\/}

\bibitem{rubio2024accurate}
Rubio-Garc{\'\i}a {\'A}, Fern{\'a}ndez-Lorenzo S, Garc{\'\i}a-Ripoll J~J and Porras D 2024 {\em Physica A: Statistical Mechanics and its Applications\/} {\bf 639} 129637

\bibitem{hosmer2013applied}
Hosmer~Jr D~W, Lemeshow S and Sturdivant R~X 2013 {\em Applied logistic regression\/} (John Wiley \& Sons)

\bibitem{fisher1936use}
Fisher R~A 1936 {\em Annals of eugenics\/} {\bf 7} 179--188

\bibitem{theodoridis2015machine}
Theodoridis S 2015 {\em Machine learning: a Bayesian and optimization perspective\/} (Academic press)

\bibitem{nocedal1999numerical}
Nocedal J and Wright S~J 1999 {\em Numerical optimization\/} (Springer)

\bibitem{fletcher2000practical}
Fletcher R 2000 {\em Practical methods of optimization\/} (John Wiley \& Sons)

\bibitem{parlett1998symmetric}
Parlett B~N 1998 {\em The symmetric eigenvalue problem\/} (SIAM)

\bibitem{horn2012matrix}
Horn R~A and Johnson C~R 2012 {\em Matrix analysis\/} (Cambridge university press)

\bibitem{robert1999monte}
Robert C~P, Casella G and Casella G 1999 {\em Monte Carlo statistical methods\/} vol~2 (Springer)

\bibitem{mitra1986convergence}
Mitra D, Romeo F and Sangiovanni-Vincentelli A 1986 {\em Advances in applied probability\/} {\bf 18} 747--771

\bibitem{james2013introduction}
James G, Witten D, Hastie T, Tibshirani R {\em et~al.\/} 2013 {\em An introduction to statistical learning\/} vol 112 (Springer)

\bibitem{fawcett2006introduction}
Fawcett T 2006 {\em Pattern recognition letters\/} {\bf 27} 861--874

\bibitem{tibshirani1996regression}
Tibshirani R 1996 {\em Journal of the Royal Statistical Society Series B: Statistical Methodology\/} {\bf 58} 267--288

\bibitem{scikit-learn}
Pedregosa F, Varoquaux G, Gramfort A, Michel V, Thirion B, Grisel O, Blondel M, Prettenhofer P, Weiss R, Dubourg V, Vanderplas J, Passos A, Cournapeau D, Brucher M, Perrot M and Duchesnay E 2011 {\em Journal of Machine Learning Research\/} {\bf 12} 2825--2830

\bibitem{kelly2023uci}
Kelly M, Longjohn R and Nottingham K 2023 {The UCI Machine Learning Repository}

\bibitem{cios2001spectf}
Cios K, Kurgan L and Goodenday L 2001 {SPECTF Heart [Dataset]} {UCI Machine Learning Repository}

\bibitem{wolberg1993breast}
Wolberg W, Mangasarian O, Street N and Street W 1993 {Breast Cancer Wisconsin (Diagnostic) [Dataset]} {UCI Machine Learning Repository}

\bibitem{zhang2008ozone}
Zhang K, Fan W and Yuan X 2008 {Ozone Level Detection [Dataset]} {UCI Machine Learning Repository}

\bibitem{hopkins1999spambase}
Hopkins M, Reeber E, Forman G and Suermondt J 1999 {Spambase [Dataset]} {UCI Machine Learning Repository}

\bibitem{realinho2021predict}
Realinho V, Vieira~Martins M, Machado J and Baptista L 2021 {Predict Students' Dropout and Academic Success [Dataset]} {UCI Machine Learning Repository}

\bibitem{uci2020taiwanese}
 2020 {Taiwanese Bankruptcy Prediction [Dataset]} {UCI Machine Learning Repository}

\bibitem{guyon2004madelon}
Guyon I 2004 {Madelon [Dataset]} {UCI Machine Learning Repository}

\bibitem{yeh2009default}
Yeh I 2009 {Default of Credit Card Clients [Dataset]} {UCI Machine Learning Repository}

\bibitem{yasser2022loan}
Yasser~H M 2022 {Loan Default [Dataset]}

\bibitem{miasojedow2013adaptive}
Miasojedow B, Moulines E and Vihola M 2013 {\em Journal of Computational and Graphical Statistics\/} {\bf 22} 649--664

\bibitem{barash2017gpu}
Barash L~Y, Weigel M, Borovsk{\`y} M, Janke W and Shchur L~N 2017 {\em Computer Physics Communications\/} {\bf 220} 341--350

\end{thebibliography}

\end{document}